# Siamese Basis Function Networks for Data-efficient Defect Classification in Technical Domains

Tobias Schlagenhauf*

Karlsruhe Institute of Technology
wbk Institute of Production Science

Faruk Yildirim

Karlsruhe Institute of Technology
wbk Institute of Production Science

Benedikt Brückner

Karlsruhe Institute of Technology
wbk Institute of Production Science

*Abstract* — **Training deep learning models in technical domains is often accompanied by the challenge that although the task is clear, insufficient data for training is available. In this work, we propose a novel approach based on the combination of Siamese networks and radial basis function networks to perform data-efficient classification without pretraining by measuring the distance between images in semantic space in a data- efficient manner. We develop the models using three technical datasets, the NEU dataset, the BSD dataset, and the TEX dataset. In addition to the technical domain, we show the general applicability to classical datasets (cifar10 and MNIST) as well. The approach is tested against state-of-the-art models (Resnet50 and Resnet101) by stepwise reduction of the number of samples available for training. The authors show that the proposed approach outperforms the state-of- the-art models in the low data regime.**

I. INTRODUCTION

The classification of objects in various domains has been gaining attention since the development of modern and powerful deep learning techniques. [1] Until recently, the human visual system had been unreached by computer algorithms. This has changed with the development of deep learning architectures.[5] Substantial successes could, for example, be achieved in the ILSVRC-ImageNet contest [2] using deep learning architectures. But deep learning architectures have also been gaining considerable success in the technical domain [3]. To make progress on the way toward autonomous systems, tools, and machines in the industrial context – but also in all other domains –, it is important to have accurate models for the classification of objects and the quality assessment of products. Additionally, to realize autonomous systems, it is important to enable them to self-describe their condition to prevent breakdowns. This is called predictive maintenance [48]. Especially in the technical domain, one is confronted with the fact that data is often rare in a sense that it is either not available in large quantities or must be generated, which in turn is very costly [52]. An example is the automatic vision-based quality inspection of products, which is often difficult since examples of rejects are rare. The same holds for the detection of failures in the context of condition monitoring. Another challenge in the technical domain is the long tail of possible classes. Firstly, the objects which are of interest in the technical domains are numerous, like for instance failures on rails [7], failures in concrete [6], failures in wood [8], failures on metallic surfaces [4], and failures on machine tools [9], to name only a few examples. Secondly, each possible and produced product could, in principle, be a possible class. The numerous possible cases make it difficult to generate large datasets by combining datasets of the same objects as can be done in "pie, house, cat, mouse, car" cases. In addition to that, the classes rarely have counterparts in the real world, which amplifies the latter argument.

The presented approach is based on the idea of comparing images in the feature domain and assigning the class of the image most similar to the questioned image. This is depicted in **Fehler! Verweisquelle konnte nicht gefunden werden.**. The main idea of this paper is based on the concept of a nearest-neighbor classifier used in a radial basis function network (RBF network) calculating a similarity measure between an input image and a query image. A drawback of the classical RBF network is its limited size [10] together with the limitation of the L2 norm in terms of describing a semantically useful similarity measure on raw image pixel data [43]. The authors implemented the basis function kernels as so-called Siamese-Kernels where a Siamese network was implemented for each kernel in the (R)BF network. By using these kernels as feature extractors, this allows for more meaningful comparisons to be made in the feature space.

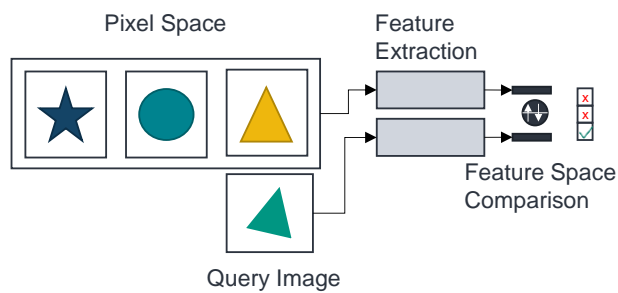

Fig. 1 Image comparison in feature space

The here presented approach is different from a classical CNN-based instance classification in that the authors perform an instance filtering using a Siamese network architecture by comparing the extracted feature vector of the query image to the extracted feature vectors of eligible images. The resulting distance is then processed in a RBF nearest-neighbor approach. The main achievements of the paper are: 1. We provide a novel deep learning architecture (SBF-Net) which classifies images based on the comparison of their semantic representations. 2. We show a novel approach to train a Siamese network with one single center to generate an ensemble of experts. 3. We show the superior data efficiency defined as validation accuracy given a specific number of training points of the SBF-Net in comparison with state-of-the-art models.

The remainder of the paper is structured as follows. Section 2 reviews the related work in the field of RBF networks and Siamese networks for failure classification together with the general approach to failure classification on metallic surfaces. Section 3 presents our own approach and discusses the approach of using Siamese networks as kernels in an RBF network. Section 4 briefly describes the representatives of the technical datasets: The Northeastern University (NEU) surface defect database [11] showing six different kinds of defects on metallic surfaces, the ball screw drive (BSD) dataset [50] of defective machine tool elements, and the fabric (TEX) dataset [51] of failures on woven textiles. The well-known MNIST and cifar10 datasets are not described further. Section 5 presents the results of the SBF-Net by first investigating the basic effect of the number of Siamese-Kernels per class as well as the effect of data on the performance of the SBF-Net using the NEU dataset. Based on these findings, the data efficiency on the NEU, KGT, TEX, MNIST, and cifar10 datasets in comparison to the classical ResNet50 and ResNet101 models is investigated by using 3,5,10,20,50, and 100 data points per class. The results are followed by a discussion. Section 6 concludes the work and states open research questions.

## II. RELATED WORK

According to [10][20], RBF networks nowadays are kind of forgotten neural network structures. Indeed, in comparison to classical CNN-based approaches, there is only a limited number of RBF-based image classification approaches such as those described, for instance, by [12] and [13]. This is likely due to the fact that the classical L2 norm is not a proper distance function to be used when dealing with raw image values in high-dimensional pixel space. Further, the classical RBF approach can be described as a version of a k-nearest-neighbor classification algorithm [10] which, for instance, [14] has proven to be underperforming in comparison to other classical machine learning algorithms like support vector machines. Nevertheless, one outstanding architecture implementing RBF elements is the well-known LeNet5 [15] architecture which uses Gaussian kernels in one of its last layers. Besides that, another recent approach described in [16] presents a deep RBF learning algorithm based on the well-known LeNet5 architecture for classification of the MNIST dataset. In [17], for instance, one finds a somewhat earlier application of RBF networks which use an RBF network for the classification of texture images. There, the authors emphasize the relevance of correctly choosing the prototype centers. Another contemporary application of RBF networks presented in [18] is the use of an RBF module as part of a pipeline for breast cancer detection in medical images. The images, though, are not processed in their raw formats. Yet another application of RBF networks in the medical image classification sector can be found in [19], where the authors used an RBF network for the classification of brain diseases by extracting classification features in advance. An interesting application is the use of RBF networks in the work by [20]. Here, the ReLu activation function in classical convolutional neural networks is replaced by RBF kernels to classify the MNIST, cifar10, and cifar100 datasets. It was found that using RBF activation functions is difficult since the network easily gets stuck in local minima during training.

Siamese networks have recently experienced significant attention due to their successful application in numerous domains. Substantial progress, enabled by their use, was made in the field of computer vision, especially in face recognition applications [21]. Nevertheless, their potential extends to other fields of research as well, e.g. to natural language processing [22] and object tracking [23].

The use of Siamese architectures for the purpose of defect detection, as showcased in this work, is an area of research that has been studied only insufficiently. Few works explore the potential of these approaches, but the results obtained are generally promising. [24] demonstrates that once trained on a specific task, such network may easily be reused for different purposes. Particular cases of application are presented in [25], where defective buttons are identified using Siamese networks, and [26], where the quality of a steel is assessed based on the appearance of its surface.

On the contrary, defect classification approaches leveraging different architectures are considered more frequently. They are mostly used to detect faults appearing on the surface of steel, and there is a broad variety of models for this purpose. [27] presents a detector using shearlet encoding and linear regression, while [28] models defect classes using hyperspheres in order to recognize potential surface anomalies. Further approaches include [29], employing kernel classifiers for detection, and [30], where a CNN network is used. Use of convolutional networks for this application is quite common. Another example may be found in [31] for more general applications beyond steel inspection. Lastly, [32] describes a system that learns through inputs provided by an expert. Various works demonstrate that these techniques can be used for other materials, too. Cracks in electrical components are detected in [33] through image segmentation performed by CNNs. A problem that appears more difficult to the human eye is the treatment of fabric due to its irregular surface. However, even such difficult problems may be solved as is demonstrated in [34] using autoencoders.

One will easily notice that most of the aforementioned approaches employ deep learning techniques in order to outperform earlier models. This is part of a greater trend that may be observed in various fields of research [35]. Besides the works addressing the topic of defect classification in general, many research projects have specifically investigated the NEU dataset which we use as one technical dataset to demonstrate the advantage of our approach. [36] generates features using a CNN variant and then classifies the NEU images using a heuristic. Convolutional networks are equally employed in [37], the features generated are then fed into a fusion and a region proposal network before classification takes place. A major drawback of CNN approaches is the fact that training the network is usually expensive in terms of time and resources. To tackle this issue, [38] implements a transfer learning approach using pretrained networks and obtains promising results. [39] proposes a classifier which, once trained, may easily be adapted to changing conditions, such as an alteration of the production process supervised by the model. An approach particularly robust to noisy inputs, which are likely to occur in a real-world setting, is presented in [40].
*Summa summarum*, the literature shows that Siamese networks can serve as powerful feature extractors. Further RBF nets perform a distance-based classification but lack the fact that the pixel space is too high dimensional to achieve good results.
The here presented approach picks up the fact that the need for large datasets in the technical domain is often described in the literature but to the best of our knowledge, no investigations in terms of data efficiency in comparison to state-of-the-art models in the technical domain have been undertaken. Especially the distance-based classification using semantic feature vectors implementing the here presented architecture is new. The findings provide a novel method for both researchers and practitioners to further develop data-efficient classification algorithms in the technical domain.

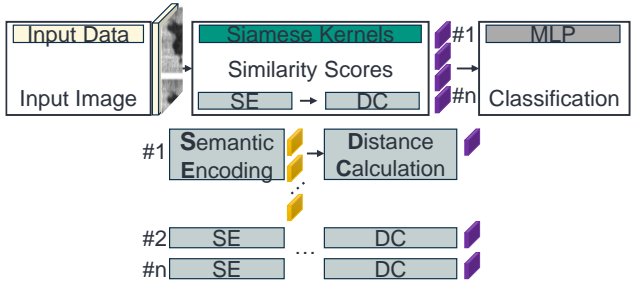

Fig. 2 Qualitative architecture of the SBF-Net

### III. OWN APPROACH

The proposed architecture is based on three main components which are depicted, combined as a Siamese basis function network (SBF-Net), in Fig. 2. The first component is the architecture of a radial basis function network as a method of performing a classification based on the comparison of samples through a similarity measure like in nearest-neighbor classification. In these networks, the distance metrics to calculate the similarity score are classical distance metrics like the Euclidean distance or the cosine similarity which are, as described by e.g. [43], an insufficient way to compute a similarity measure in high-dimensional data like image data. During classification, normally statements such as: "Are the objects shown in the image the same as those in an image of a certain class, respectively is the underlying semantics in the images the same?" are derived. Our approach is less interested in the rare differences of pixel values but in encoding the semantics in images and in obtaining similarity scores between the encoded semantics. To achieve this behavior, the authors build upon Siamese networks and use them as effective semantic-feature extractors for classification. The authors name these feature extraction units Siamese-Kernels and implement them instead of the classical radial basis function kernels in the RBF network as the first part of the SBF-Net. To reinforce the classification ability, the authors additionally equip the network with a multilayer perceptron (MLP) instead of the single-layer neural network used in classical RBF networks. In the following paragraphs, the authors will explain in detail the single components together with the training setup.

### A. Basis Function Network

The idea of a radial basis function network as proposed by [41] is to use prototype vectors to realize a weighted comparison to an input vector. In contrast to a classical neural network in which the output per node is calculated as $o_p = \sigma(\sum_k x_k w_p^k)$, a radial basis function network implements so-called RBF kernels, where the input $\boldsymbol{x}$ is compared to a prototype vector $\boldsymbol{\mu}$ which can be viewed as a class center in a nearest-neighbor approach. A distance score like the Euclidean distance is calculated on $\boldsymbol{x}$ and $\boldsymbol{\mu}$ followed by a Gaussian mapping. The output may be calculated as

$o_p = exp(-\frac{\sqrt{\sum_k (x_k - \mu_k^p)^2}}{2\sigma_p^2})$ with $\boldsymbol{\mu}_k^p$ the $k^{th}$ prototype vector of the $p^{th}$ class. The output of the whole classical RBF network is then calculated as follows: $y = \sum_p (exp(-\frac{\sqrt{\sum_k (x_k - \mu_k^p)^2}}{2\sigma_p^2}) w_{out}^p) = \boldsymbol{ow}_{out}^T$. Classification is then performed using a classical sigmoid function. The classical RBF network setup is depicted in Fig. 3. Both $\boldsymbol{\mu}$ and $\boldsymbol{x}$ are later presented as preprocessed feature vectors.

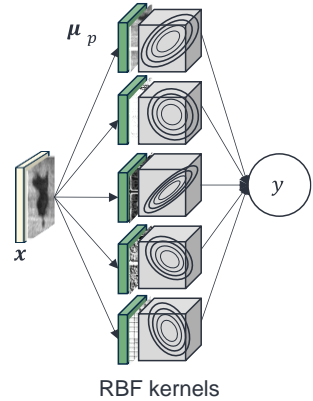

Fig. 3 Classical RBF network setup where an image $\boldsymbol{x}$ is compared to multiple prototype vectors $\boldsymbol{\mu}$ using RBF kernels which are then further processed using a weighted sum of the distances.

Due to the idea of calculating a weighted sum of multiple similarity scores to classify an input image, the RBF network is designated by e.g. [12] as an expansion of a k-nearest-neighbor classifier. The presented approach emphasizes the elimination of the well-known disadvantages of using classical distance metrics to calculate distances directly in high-dimensional space, such as pixel space in image data. To further increase the classification ability of a classical RBF network, the authors reinforced the structure by an MLP instead of a classical one-layer neural net as depicted in Fig. 3. The setup is explained later on. Unlike in the case of the classical setup, the authors use the cosine distance instead of the Euclidean distance in their calculations. The nodes in the radial basis function network are replaced by so-called Siamese-Kernels to build the architecture of the SBF-Net. These kernels are explained in the next section.

## B. *Siamese Kernels*

The basis for the Siamese-Kernels is the architecture of a Siamese network. A Siamese network, as proposed by [42], is a convolutional neural network (CNN) architecture consisting of two identical convolutional neural networks sharing weights. Using the triplet loss function, which we will explain later on, the Siamese network is trained in such a way that the distances between generated vectors are small for instances with the same class and large for instances with different classes. *Cum grano salis*, the triplet loss function is the main difference in comparison to a classical CNN architecture and led the Siamese network to produce semantically useful feature embeddings of the input instances. Since the network computes feature vectors

Here, $a$ is called an anchor, which in this case is an image of a specific class. $p$ is called positive, which is an image of the same class as the anchor, and $n$ is called negative, which is an image of a different class. $\varphi(.)$ represents the feature extractor in the form of the Siamese-Kernel. As distance function $g$, the authors implemented the cosine distance. The distance is calculated with: $ine\ Distance = 1 - \frac{x^T y}{\|x\| \times \|y\|}$, where the latter part of the equation is the classical cosine similarity. Since the model only forwards positive values, the cosine similarity takes values between 0 and 1, where 1 means that the vectors coincide. Therefore, the cosine distances take only positive numbers between 0 and 1, where values closer to 0 indicate to classify instances, these feature vectors must encode the useful semantic information needed for classification. Finally, after training, a Siamese network has learned to distinguish images based on a distance metric computed on the semantic encodings. The architecture of a Siamese neural network implementing the VGG16 CNN [44] is depicted in Fig. 5. A version of this architecture is used as basis for the Siamese-Kernels in the SBF-Net to map $\boldsymbol{x}$ and $\boldsymbol{\mu}$ from pixel space to semantic space. The single components of the Siamese-Kernels are explained together with the training setup in the following.

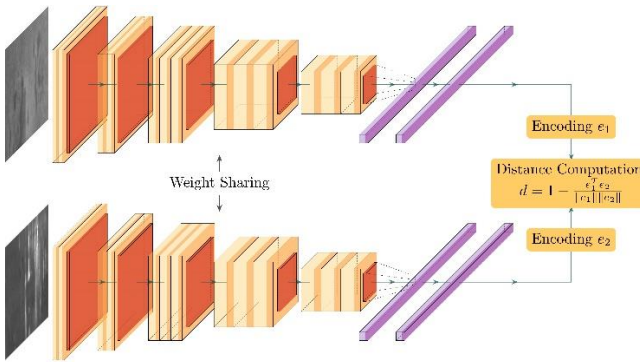

Fig. 5 Implemented Siamese network architecture based on the VGG16 backbone which shares weights (identical VGG16 models) between the branches of the Siamese network [49]

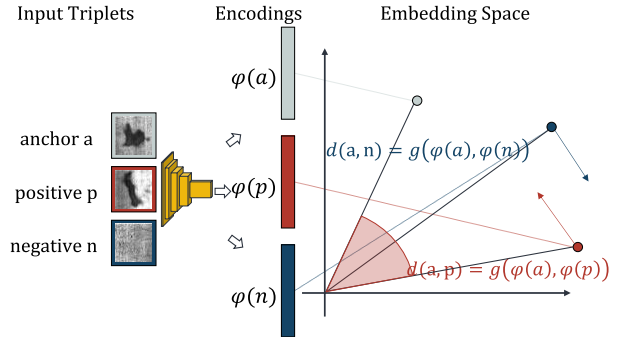

Fig. 4 Triplet loss implementing and visualizing the cosine distance in 2D where $\boldsymbol{\varphi}(.)$ stands for the encoding Siamese network.

**Convolutional Neural Network:** The authors implemented a CNN based on the VGG16 architecture with the difference that the authors use 100 instead of 4096 neurons in the fully connected layers to make the feature representation more dense. The network starts in the first layer with a feature map of size 200×200×64 and ends with size 6×6×512 in the last convolutional layer. The feature matrix is flattened and fed into two fully connected layers with 100 neurons each. ReLu is used as activation function in all models in all layers.

The output of the second fully connected layer is used as a feature vector for the following distance computations. A single Siamese network is trained with the Adam optimizer and triplet loss as loss function [45]. The triplet loss forms a core element of the Siamese network architecture and can be formalized as:
$Triplet\ Loss = \max(0, g(\varphi(a), \varphi(p)) + \alpha - g(\varphi(a), \varphi(n)))$

larger similarity (or lower distance) and values closer to 1 indicate smaller similarity (or larger distance). $\alpha$ is a so-called margin parameter to ensure encodings where the distance between the anchor and the negative is larger than the distance between the anchor and the positive but smaller than the distance between the anchor and the positive plus some margin. The vectors for the triplet loss are 100 dimensional vectors. The idea of the triplet loss together with the cosine distance is presented in Fig. 4.

Using this setup, the Siamese-Kernels learn to distinguish between images of the same class and images of different classes by pushing away images from different classes and pulling images which share classes. After training, the Siamese-Kernels return small values for images which belong to the same class and large values for images which belong to different classes. In the presented approach, the kernels are trained with a learning rate of 0.00001, for 5000 iterations using semi-hard triplet loss with margin $\alpha$ of 0.3.

**Siamese Network Implementation:**

A key aspect of the approach is the specific implementation of the Siamese-Kernels and their subsequent combination. The kernel networks are trained with one constant anchor per network. In this setup, the authors pick one image to be used as anchor of the Siamese network in advance and train the network randomly drawing positives and negatives from the training set using the triplet loss function. The anchor remains constant during training. Therewith, the center image is compared to different setups of positive and negative images and learns to accurately classify the center image as belonging to one specific class. If an image of the same class is presented after training, the network ideally returns a feature vector which results in a small distance value, whilst it returns a large distance value for images of other classes. Prior to the training, $k$ center images are selected randomly for each class $c$ in the dataset $D$ which leads to a total number of $|C|*k$ kernel networks, each one specialized on distinguishing the class of its anchor. This aspect is important because it prevents the model from overfitting to the specific center images but creating a kind of class awareness over the ensemble of kernels. This approach could be considered as an association of individual experts who make a joint decision

Using a fixed center per kernel, the classical classification task is kind of reversed since it is not the image which is assigned a class label, but it is rather the network that tells if the center (prototype) belongs to the same class as the input image. Since each Siamese-Kernel distinguishes all classes from the respective prototype class by comparing input encodings to prototype encodings, the prototypes are by design the ideal centers for the radial basis function computation. To yield a Siamese-Kernel, a Siamese net is implemented together with a Gaussian mapping in feature space.

### C. *Siamese Basis Function Network (SBF-Net)*

Using the described approach, the Siamese networks allow effective encoding of the image information for classification. The architecture of the SBF-Net is now built by combining the single Siamese networks with the RBF structure by replacing Fig. 3. with Siamese-Kernels which then preprocess the $x$ from pixel to feature space. The $\mu$ are chosen as center images from specific classes. The whole structure is depicted in Fig. 6.

Each of the Siamese kernels returns a similarity value

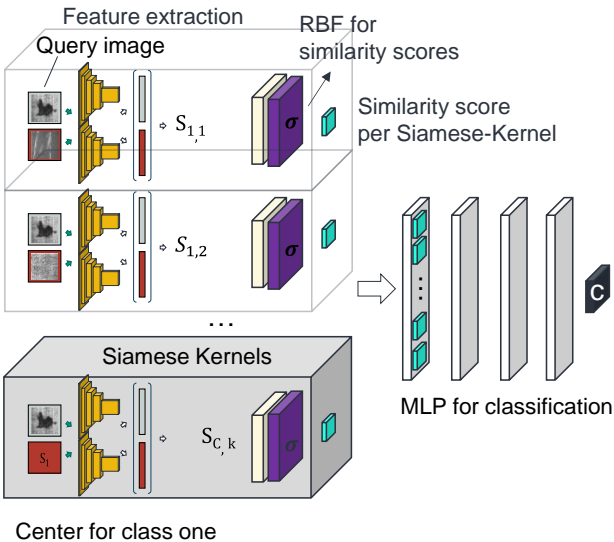

Fig. 6 SBF-Net architecture which extracts features from multiple centers per class and a query image and compares them using a RBF function before MLP classification

measuring the semantic distance between the center image and the image presented as the input. Note that each input image is compared to each Siamese node, hence the respective Siamese network in the kernels returns feature vectors which encode the affiliation between the input image and the images used as centers. Since different images are used as centers, it is more likely that similarities between images are discovered. For each image at the input, a $|C|*k$-dimensional feature similarity vector is output. This vector encodes the similarity information between the input image and the centers, which is then passed into an MLP for classification. To increase the classification ability, the authors implemented a four-layer neural network with 50 neurons per layer implementing ReLU activations. The authors implemented dropouts with a value of 0.1 after each layer. We appended an output layer with softmax activations for multiclass classification and trained the network with categorical cross entropy loss for 1000 iterations with a learning rate of $10^{-6}$.

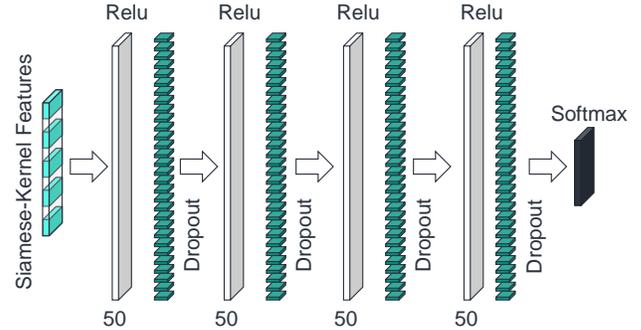

Fig. 7 MLP classifier for the Siamese-Kernel features processed by a RBF function

## IV. DATASETS

As datasets of interest, the authors chose three technical and two non-technical datasets. As representatives of the non-technical datasets, the authors chose the well-known MNIST and cifar10 datasets which will not be further described here. As first representative of the technical datasets, the authors used the Northeastern University (NEU) surface defect database [11]. The NEU dataset is a state-of-the-art dataset for defect classification and detection on metallic surfaces. It depicts six kinds of common defects on metallic surfaces of hot-rolled steel strips: rolled-in scale, patches, crazing, pitted surface, inclusion, and scratches. The dataset consists of 1800 grayscale images, where each image is 200x200 pixels. The images are equally split by categories (300 each). The classes have a low inter-class variance while the intra-class variance is high. Additionally, there are different lighting conditions, which altogether leads to situations of similar-looking images between classes, which complicates the classification. Examples of the technical datasets are shown in Fig. 8. For training, we split the images randomly in 80% for training and 20% for testing.

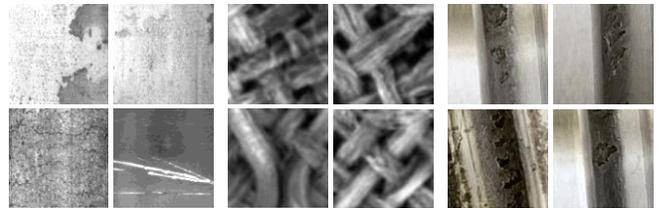

Fig. 8 NEU surface defect, TEX and BSD datasets

The second representative of the technical datasets is the TEX dataset [51] (originally called fabric dataset) which shows five different types of failures in textiles together with one "good" class. The failures are color, cut, hole, thread, and metallic contamination. The dataset contains of 108.000 64x64 pixel grayscale images where the classes are equally represented. As can be seen in the example images in Fig. 8, the defects are neither easy to distinguish nor is it trivial to specify a failure at all – at least for the human inspector. It will be interesting to see how the model can generalize using only a limited number of samples to learn from. The third technical dataset (BSD [50]) is a dataset showing failures on ball screw drives (BSD). Ball screw drives are important machine tool elements installed in most industrial machines, and an unforeseen defect can lead to unwanted idle times with severe influences on the overall equipment effectiveness (OEE). Hence, it is important to find defects on the BSD as early as possible. The dataset is made up of two classes represented by images showing a defect and images not showing a defect. The dataset contains of 21.835 150x150 pixel RGB images scaled to 100x100 pixels by the authors. The dataset contains of 11.075 without defect and 10.760 with defect and hence is nearly equally split. The dataset contains edge images where the defect is covered by soil or other pollutions and it is even difficult for the human domain

0.03*255, Random multiplication of the channel values with a value between 0.7 and 1.3 as well as a perspective transformation within a scale of (0, 0.15). All images are normalized to values between 0 and 1. The validation images are not augmented.

V. EXPERIMENTS & RESULTS

In the section below, the experiments are described followed by the associated results as well as the discussion of the results. The focus of the experiments lies on the classification of technical datasets (NEU, KGT, TEX). The experiments on the cifar10 and MNIST datasets can be viewed as an ablation study which should show the performance and transferability of the approach to non-technical domains.

The results are structured in three main research blocks. 1. The development of the SBF-Net architecture which has been shown above. 2. The effect of the number of kernels per class as well as the amount of data available for model training, and 3. The performance and data efficiency of the SBF-Net in comparison with state-of-the-art models.

A critical aspect of the SBF-Net is the number of kernels used per class. The hypothesis is that the performance of the model increases with an increasing number of kernels per class. Fig. 9 expert to label the images correctly. Since the defects occur in different sizes, the transition from images with no defect to images showing a defect is continuous especially when pollution comes into play.

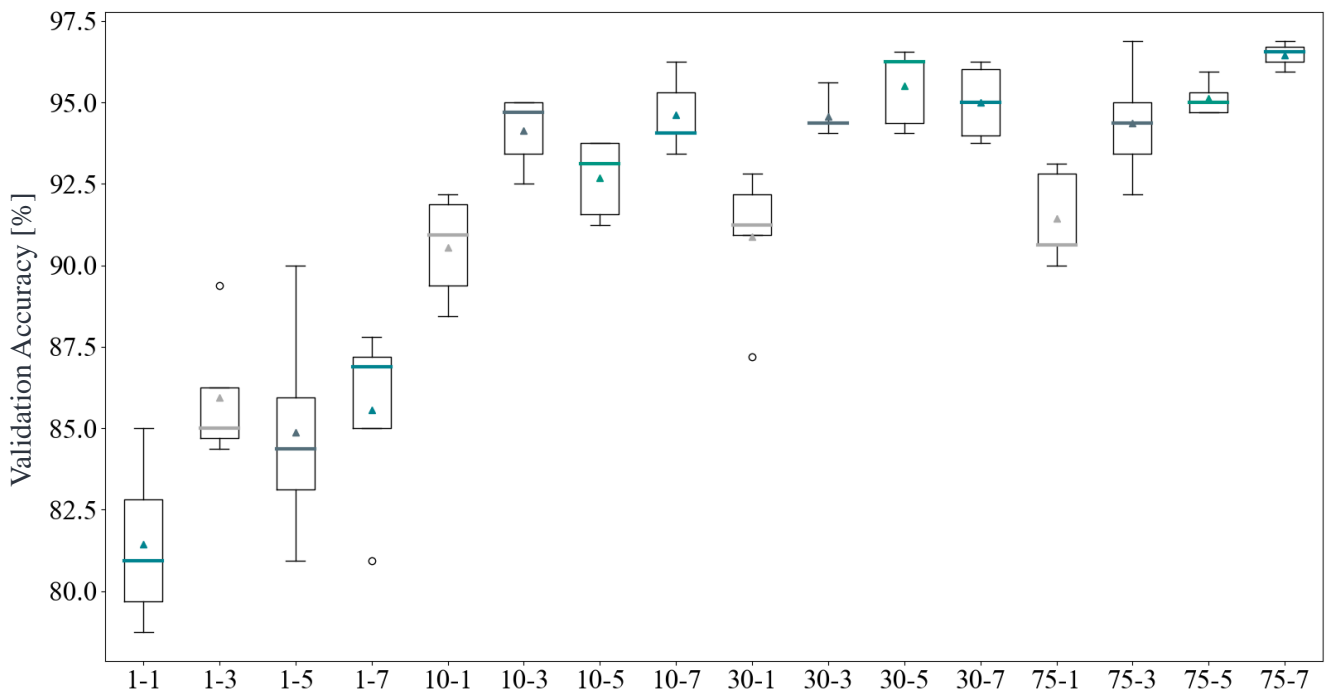

Fig. 9 Model performance as the number of training data points and centers per class increases. Structure of x-axis labeling is [Percentage-Centers]

Each image in the training datasets is used in its original form together with a four times random augmentation with the following imgaug classes: All channels contrast limited histogram equalization (CLAHE) with clip limit of (1, 10), random rotation between +-5°, 30% chance of horizontal and vertical flipping, Laplace noise with a per-channel scale of depicts the validation accuracy on the KGT dataset when training the SBF-Net with altering training data sizes of 0.1%, 1%, 10%, 30%, and 75% and an altering number of kernels per class (1,3,5,7). In the experiments, always 25% of the KGT dataset was set aside as test set.

It can be seen that the performance of the model increases with increasing size of the dataset. However, the increasing rate flattens out towards larger datasets which is a well-known effect in training deep learning models. Considering the accuracy with altering numbers of centers, the hypothesis was that the performance of the model increases with an increasing number of centers per class. Overall, this effect can be confirmed but there is a large variation over the number of centers. The interpretation of this effect is that the images for the centers of the single kernels are chosen randomly from the dataset. This results in selections which are more representative of the given task and selections which are less appropriate. Hence, the choice of the centers seems to have a significant effect on the performance of the model which can lead to situations in which 3 centers perform better than 5 or 7 centers per class. Given a large enough dataset, an open research question which could be possibly addressed by active learning strategies like [53] is how to choose centers which are optimal. However, this is only applicable given a large enough dataset. With small datasets, this effect can probably be neglected since there are no options to choose the data. Given the fact that, given enough data, this choice can be done for any possible number of centers, the authors choose 5 centers per class for the further experiments. We will see that this suffices for showing the validity of the approach.

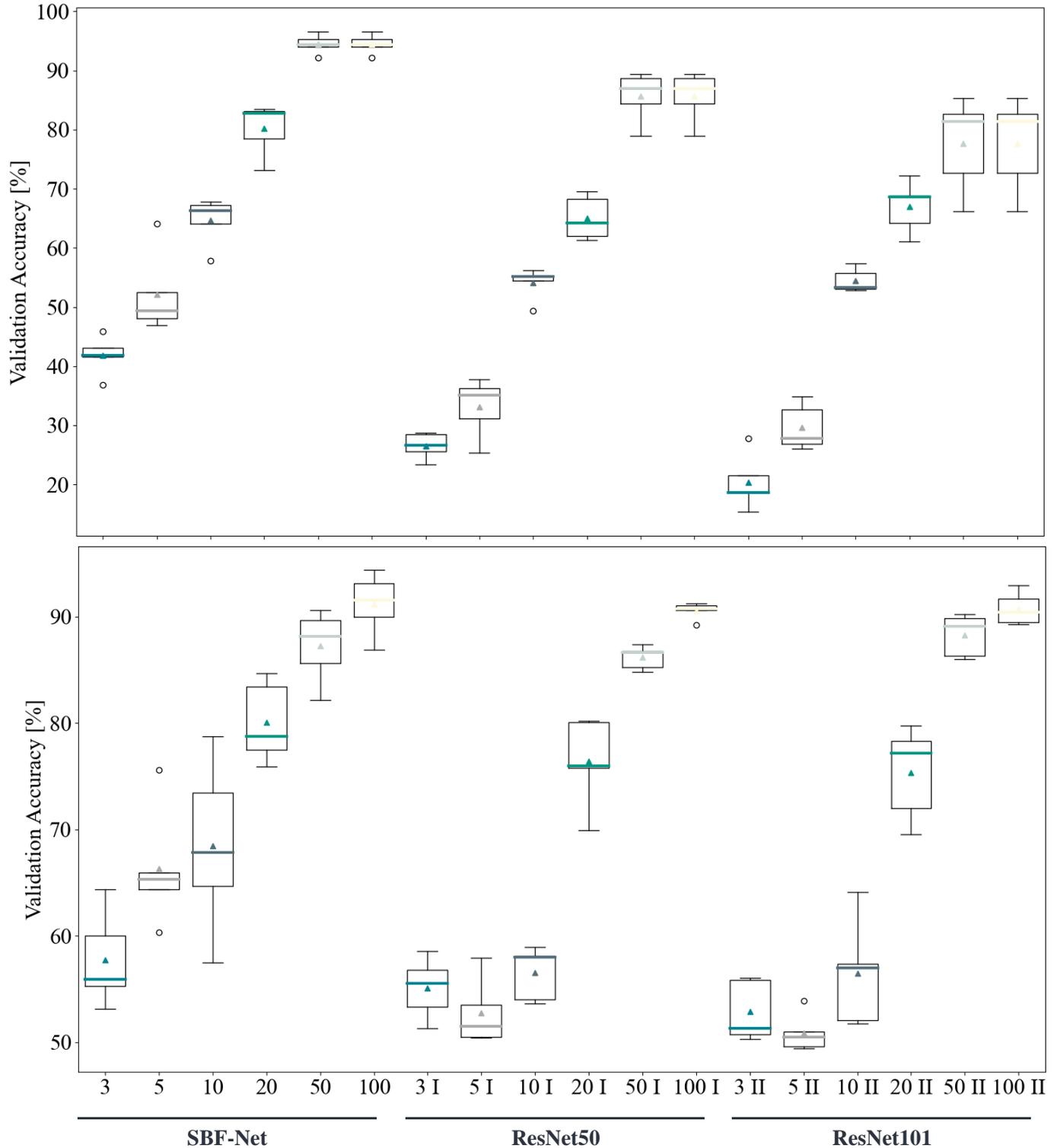

Fig. 10 Data efficiency of the SBF-Net against ReNet50 and ResNet101 for the NEU and KGT datasets

and other regular basic structures whereas the cifar10 dataset is very diverse. Since the ResNet50 and ResNet101 models have more learning capacity because of their size, they may be better suited to fit the diverse data. The single kernels in the SBF-Net may not have the performance to encode the differences or similarities between the images given as centers and all other

**Table 1** Data efficiency of the SBF-Net against ReNet50 and ResNet101 for the TEX, MNIST, and CIFAR10 Datasets

| SBF-NET | | | | | | ResNet50 | | | | | | ResNet101 | | | | | |
|---|---|---|---|---|---|---|---|---|---|---|---|---|---|---|---|---|---|
| 3 | 5 | 10 | 20 | 50 | 100 | 3 | 5 | 10 | 20 | 50 | 100 | 3 | 5 | 10 | 20 | 50 | 100 |
| TEX | | | | | | | | | | | | | | | | | |
| 0.209 | 0.269 | 0.334 | 0.294 | 0.466 | 0.491 | 0.249 | 0.210 | 0.259 | 0.321 | 0.357 | 0.490 | 0.177 | 0.198 | 0.276 | 0.305 | 0.346 | 0.487 |
| 0.191 | 0.247 | 0.231 | 0.275 | 0.387 | 0.512 | 0.233 | 0.234 | 0.257 | 0.293 | 0.368 | 0.451 | 0.235 | 0.176 | 0.263 | 0.320 | 0.378 | 0.470 |
| 0.275 | 0.269 | 0.319 | 0.316 | 0.416 | 0.469 | 0.177 | 0.236 | 0.276 | 0.318 | 0.361 | 0.464 | 0.234 | 0.233 | 0.291 | 0.311 | 0.361 | 0.443 |
| 0.262 | 0.237 | 0.278 | 0.303 | 0.369 | 0.484 | 0.247 | 0.216 | 0.270 | 0.303 | 0.368 | 0.440 | 0.178 | 0.203 | 0.279 | 0.295 | 0.371 | 0.417 |
| 0.247 | 0.284 | 0.319 | 0.303 | 0.434 | 0.509 | 0.191 | 0.252 | 0.251 | 0.281 | 0.400 | 0.477 | 0.180 | 0.207 | 0.310 | 0.287 | 0.368 | 0.447 |
| 0.237 | 0.261 | 0.296 | 0.298 | 0.414 | 0.493 | 0.219 | 0.230 | 0.263 | 0.303 | 0.371 | 0.464 | 0.201 | 0.204 | 0.284 | 0.304 | 0.365 | 0.453 |
| MNIST | | | | | | | | | | | | | | | | | |
| 0.691 | 0.775 | 0.881 | 0.950 | 0.959 | 0.981 | 0.172 | 0.486 | 0.766 | 0.872 | 0.926 | 0.954 | 0.112 | 0.272 | 0.507 | 0.782 | 0.876 | 0.936 |
| 0.753 | 0.791 | 0.903 | 0.913 | 0.959 | 0.972 | 0.252 | 0.307 | 0.832 | 0.880 | 0.916 | 0.954 | 0.198 | 0.499 | 0.675 | 0.824 | 0.879 | 0.900 |
| 0.806 | 0.719 | 0.875 | 0.934 | 0.956 | 0.981 | 0.176 | 0.466 | 0.744 | 0.859 | 0.933 | 0.952 | 0.160 | 0.382 | 0.642 | 0.850 | 0.916 | 0.943 |
| 0.688 | 0.806 | 0.894 | 0.928 | 0.966 | 0.969 | 0.190 | 0.475 | 0.735 | 0.898 | 0.930 | 0.958 | 0.190 | 0.276 | 0.690 | 0.853 | 0.885 | 0.916 |
| 0.625 | 0.747 | 0.875 | 0.922 | 0.950 | 0.972 | 0.175 | 0.448 | 0.718 | 0.887 | 0.929 | 0.959 | 0.121 | 0.286 | 0.673 | 0.831 | 0.893 | 0.908 |
| 0.713 | 0.767 | 0.886 | 0.929 | 0.958 | 0.975 | 0.193 | 0.436 | 0.759 | 0.879 | 0.927 | 0.955 | 0.156 | 0.343 | 0.637 | 0.828 | 0.890 | 0.921 |
| CIFAR10 | | | | | | | | | | | | | | | | | |
| 0.206 | 0.219 | 0.206 | 0.213 | 0.356 | 0.259 | 0.146 | 0.188 | 0.263 | 0.269 | 0.328 | 0.398 | 0.134 | 0.197 | 0.235 | 0.253 | 0.281 | 0.275 |
| 0.169 | 0.200 | 0.259 | 0.259 | 0.297 | 0.131 | 0.138 | 0.170 | 0.254 | 0.270 | 0.329 | 0.360 | 0.114 | 0.150 | 0.215 | 0.221 | 0.313 | 0.318 |
| 0.216 | 0.203 | 0.262 | 0.269 | 0.281 | 0.259 | 0.132 | 0.173 | 0.238 | 0.277 | 0.355 | 0.385 | 0.118 | 0.152 | 0.233 | 0.262 | 0.331 | 0.243 |
| 0.188 | 0.206 | 0.253 | 0.228 | 0.275 | 0.309 | 0.162 | 0.180 | 0.274 | 0.269 | 0.339 | 0.394 | 0.132 | 0.162 | 0.222 | 0.271 | 0.316 | 0.262 |
| 0.178 | 0.172 | 0.219 | 0.247 | 0.188 | 0.228 | 0.121 | 0.189 | 0.243 | 0.281 | 0.315 | 0.335 | 0.122 | 0.170 | 0.220 | 0.267 | 0.287 | 0.297 |
| 0.191 | 0.200 | 0.240 | 0.243 | 0.279 | 0.238 | 0.140 | 0.180 | 0.254 | 0.273 | 0.333 | 0.375 | 0.124 | 0.166 | 0.225 | 0.255 | 0.306 | 0.279 |

Using 5 centers per class, the authors trained the SBF-Net on the KGT, NEU, TEX, cifar10, and MNIST datasets with an increasing number of data points (3,5,10,20,50,100) using 25% of the data as hold out test set. The results for the NEU and KGT datasets are shown in Fig. 10 while the results for the TEX, MNIST, and cifar10 datasets are shown in Table 1. Considering the results for the classification of the NEU and KGT data, the SBF-Net performs about 10% better than the ResNet50 and ResNet101 for 3,5, and 10 images per class. Using 20 data points per class, the advance of the SBF-Net decreases for the KGT data and stays constant for the NEU data. For 50 and 100 data points per class, the SBF-Net is on a par or slightly better than the state-of-the-art models. This picture is also reflected in Table 1. For 3,5, and 10 data points, the SBF-Net is leading. Whereas for 20 TEX images the performance drops, the difference in the accuracy is in principle neglectable. In the MNIST case, the model is leading for all data set sizes. The picture is a little bit more diverse in the CIFAR10 case. Here, the model is still the best for 3 and 5 data points but then gets passed by the other models. This could be explained by the fact that the CIFAR10 dataset is structurally different from the TEX, MNIST, KGT, and NEU datasets. These datasets could be described as sharing some prominent features like lines, edges, images in the dataset. To get a better picture of the overall performance, Fig. 11 shows the average validation accuracy over all datasets. The large advantage of the SBF-Net especially in the low data regime is obvious. But also for the larger data setups, the model performs on par with the state-of- the-art models. *Summa summarum*, it could be concluded that the SBF-Net architecture has an advantage when it comes to low data sizes and can even be as accurate as the state-of-the-art models for larger datasets. A remarkable result is achieved when training the model on 75% of the NEU data as shown in Fig. 11. The achieved 99.69 percent are above the current state of the art. As supporting argument, the benchmark in [4] contains several versions of the ResNet architecture. The performance of the classic ResNet-50 architecture used by the authors is highlighted as well.

## VI. CONCLUSION

The motivation of the work was to provide a novel data-efficient method which can classify images from the technical domain (KGT, NEU, TEX) using small amounts of training data. The goal was to show this without pretraining and transfer learning. In addition, the generality of the approach should be shown on non-technical datasets (MNIST, cifar10) as well. To achieve this goal, the authors proposed a new so-called SBF-Net model which is based on a combination of multiple Siamese networks and a radial basis function network in which the Siamese nets are used as so-called Siamese-Kernels. The model then computes semantically relevant feature vectors and performs a distance-based classification. An important aspect was the training of the Siemese-Kernels with one specific center image per kernel. The single kernels learned the semantic representation of the center images in comparison to all other images. This approach led the whole SBF-Net to some kind of class awareness. The authors showed that the proposed architecture works well in low data regimes and outperforms classical state-of-the-art models with respect to data efficiency measured by the validation accuracy for a given number of training images. The authors also showed that the SBF-Net achieves comparable results even in larger data domains and is able to outperform state-of-the-art models there. The presented approach should open a new chapter in the field of data-efficient similarity-based deep learning research.

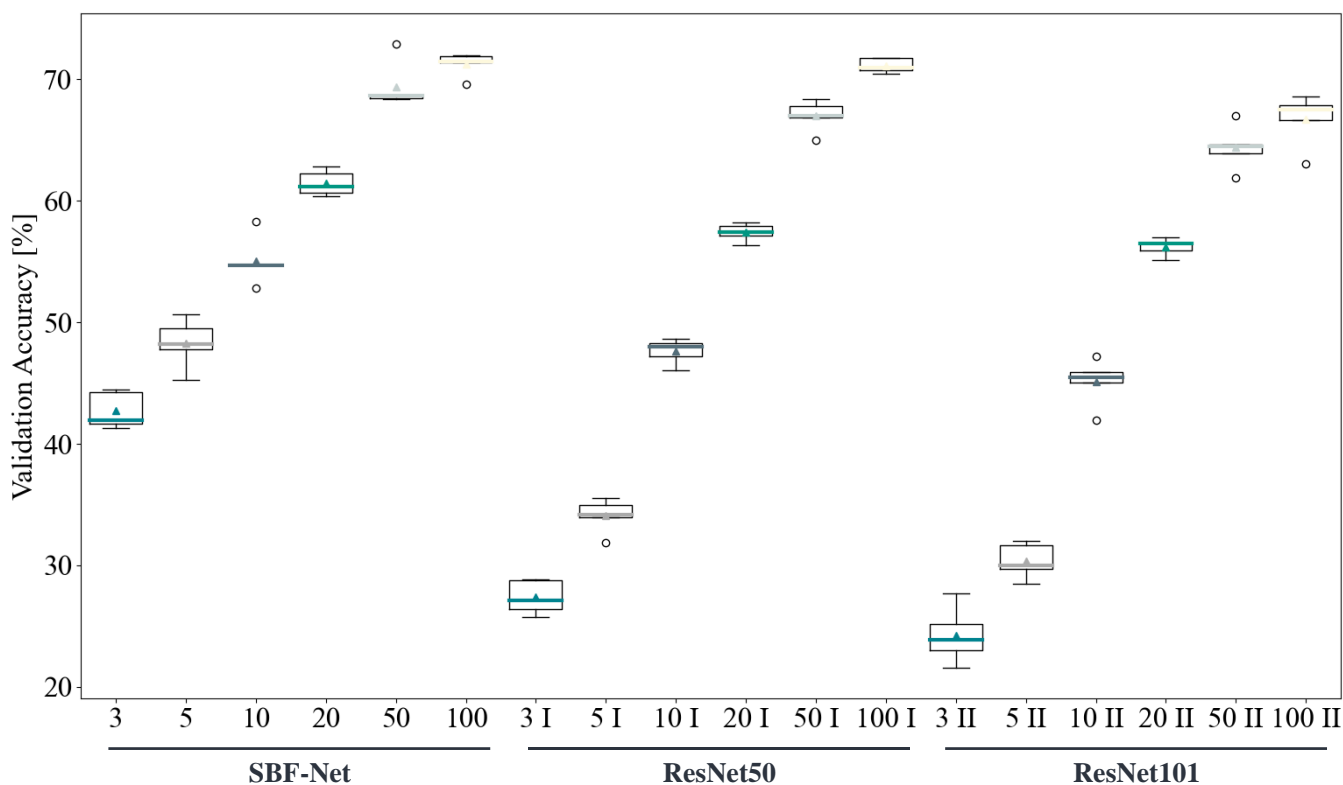

Fig. 11 Average performance of the models over all datasets given 3, 5, 10, 20, 50, and 100 data points per class

A limitation which has to be further investigated is the drop in performance for the cifar10 dataset for larger data set sizes. A hypothesis which must be further investigated is that the variation in the vectors describing the different objects in the cifar10 dataset is too large to be mapped by the model structure. A reinforcing argument could be found in the way the SBF-Net is trained using one center image per kernel. In this setup, the model must learn to find the differences and similarities between the single center image and all other positive and negative images. If now a center image is randomly chosen which is not a distinctive representative of its class, then the kernel may be kind of confused. This effect could increase with larger dataset size. A practical way to check this in further experiments is the use of much larger models.

In addition to that, comparing the performance of all three models on the cifar10 dataset in comparison to e.g. [54] who also checked the performance of a ResNet20 on 10 data points per class, it is notable that the performance is lower even though the architectures are quite similar. This could most likely be reduced to the fact that the authors of [54] designed their data augmentation strategy for the cifar10 dataset while here, the data augmentations are designed to aid a model trained on technical dataset. This could lead to insufficient augmentations in the cifar10 case which do not help but even harm the model performance. This should be checked in further experiments. Another interesting aspect already mentioned above is the choice of the center images for the Siamese-Kernels. This effect may be diminishing with smaller data sets. However, it has been shown that the proposed architecture can serve as a very strong classifier even in larger data domains and here, it is interesting what performance the model can achieve if perfect center images are chosen and the number of kernels is increased at the same time. The latter thought comes with a practical limitation which should also be investigated in further experiments. Since for each kernel, a Siamese network is trained, the needed computation increases linearly with the number of kernels. Hence, there should be developed ways to increase the number of kernels and at the same time reduce the model complexity of the single kernels such that the overall needed computation stays the same. A promising direction could be the use of knowledge distillation as described by e.g. [55]. Another question that directly emerges from the research results and needs to be further investigated is how to further increase the performance of the model even with small data sets and close the gap to the performance achieved with large data sets. Thinkable approaches are general transfer-learning approaches which does not only work for one limited domain but for a large number of domains by extracting generally applicable features which are not only located in the lower layers but are also located in deeper layers. Some kind of learnable drop-out or selection mechanism/strategy could be implemented to only use specific features when needed for a specific tasks.

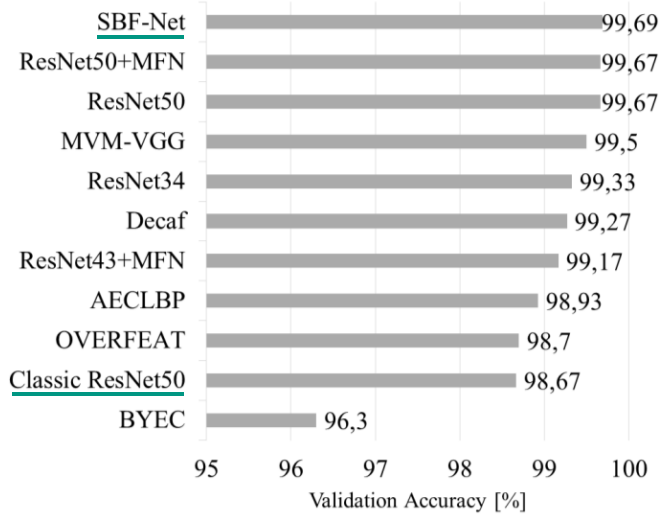

Fig. 12 Comparison of classification results based on [4]